\documentclass[conference]{IEEEtran}
\IEEEoverridecommandlockouts
\usepackage{cite}
\usepackage{amsmath,amssymb,amsfonts}
\usepackage{algorithmic}
\usepackage{graphicx}
\usepackage{textcomp}
\usepackage{xcolor}

\usepackage{times}
\usepackage{epsfig}
\usepackage{graphicx}
\usepackage{amsmath}
\usepackage{amssymb}

\usepackage{multirow}
\usepackage{float}
\usepackage{amsfonts}
\usepackage{color}
\usepackage{subcaption}
\usepackage[super]{nth}

\usepackage[pagebackref=true,breaklinks=true,letterpaper=true,colorlinks,bookmarks=false]{hyperref}

\def\BibTeX{{\rm B\kern-.05em{\sc i\kern-.025em b}\kern-.08em
    T\kern-.1667em\lower.7ex\hbox{E}\kern-.125emX}}

\def\para#1{\smallskip\noindent{\bf{#1}}}

\newcommand{\argmin}{\mathop{\rm arg~min}\limits}

\makeatletter

\def\eg{\emph{e.g}.} 
\def\ie{\emph{i.e}.}

\def\etal{\emph{et al}.~}
\makeatother

\begin{document}

\title{VIO-Aided Structure from Motion Under Challenging Environments
}

\author{\IEEEauthorblockN{Zijie Jiang\IEEEauthorrefmark{1}, Hajime Taira\IEEEauthorrefmark{1}, 
Naoyuki Miyashita\IEEEauthorrefmark{2}, Masatoshi Okutomi\IEEEauthorrefmark{1}}
\IEEEauthorblockA{\IEEEauthorrefmark{1}
Dept. of Systems and Control Engineering, Tokyo Institute of Technology\\
Email: \{zjiang, htaira, mxo\}@ok.sc.e.titech.ac.jp
}
\IEEEauthorblockA{\IEEEauthorrefmark{2}R\&D Group, Olympus Corporation\\
Email: naoyuki.miyashita@olympus.com}
}

\maketitle

\begin{abstract}
In this paper, we present a robust and efficient Structure from Motion pipeline for accurate 3D reconstruction under challenging environments by leveraging the camera pose information from a visual-inertial odometry.
Specifically, we propose a geometric verification method to filter out mismatches by considering the prior geometric configuration of candidate image pairs.
Furthermore, we introduce an efficient and scalable reconstruction approach that relies on batched image registration and robust bundle adjustment, both leveraging the reliable local odometry estimation.
Extensive experimental results show that our pipeline performs better than the state-of-the-art SfM approaches in terms of reconstruction accuracy and robustness for challenging sequential image collections.
\end{abstract}


\section{Introduction \label{sec:intro}}
\noindent
3D reconstruction with accurate geometry is desired for many different applications, such as robot navigation and industrial inspection.
Structure from Motion (SfM) is a common technique to achieve this goal, which aims to recover 3D geometry and camera poses from image collections of a target scene~\cite{frahm2010building,schonberger2016structure}.
Given well-conditioned image collections, SfM can achieve highly precise 3D reconstruction assured by accurate camera pose estimation using rich local feature correspondences~\cite{fischler1981random,lowe2004distinctive}
and subsequent global bundle adjustment~\cite{triggs1999bundle}
that refines the camera poses and structures.
However, these approaches are vulnerable to the degradation of visual information such as the absence of textures and lack of overlapping views,
consequently fail to find a good initial pose, resulting in incomplete or broken 3D structure (cf. Fig.~\ref{fig:corridor_sample}).

{\tabcolsep=1pt
\begin{figure}[t]
    \centering
    \begin{tabular}{cccccc}
    \multicolumn{2}{c}{\includegraphics[width=0.33\linewidth]{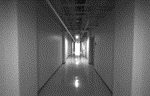}} &
    \multicolumn{2}{c}{\includegraphics[width=0.33\linewidth]{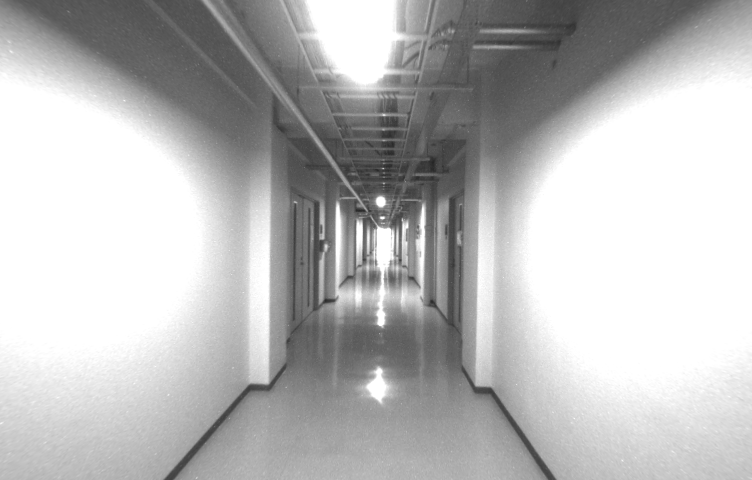}} &
    \multicolumn{2}{c}{\includegraphics[width=0.33\linewidth]{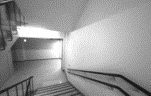}} \\[1pt]
    \multicolumn{2}{c}{\nth{2} floor} &  \multicolumn{2}{c}{\nth{3} floor} & \multicolumn{2}{c}{stairs} \\[3pt]
    \multicolumn{6}{c}{\includegraphics[width=1\linewidth]{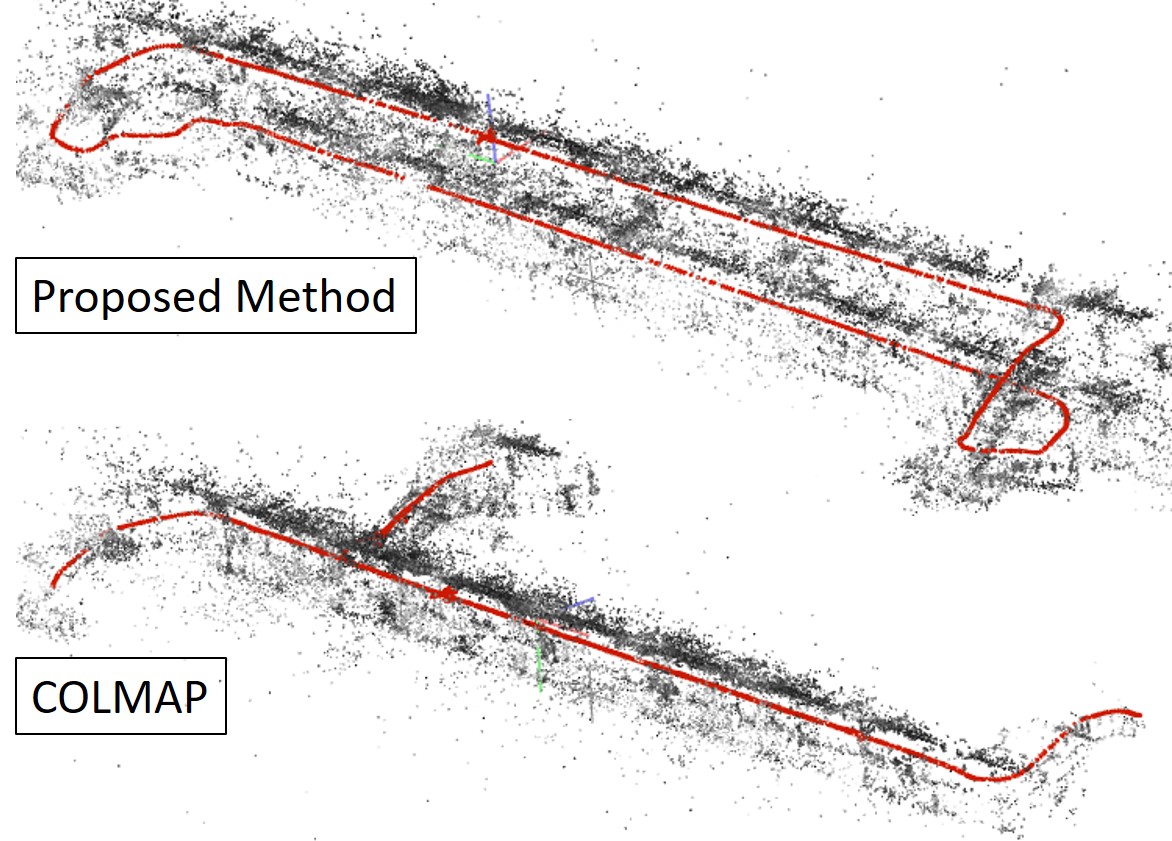}} \\[3pt]
    \end{tabular}
    \caption{{\bf 3D reconstruction results for a challenging indoor scene.} Gray dots are the reconstructed 3D scene points and red cones represent the estimated cameras relative to the model. 
    COLMAP incorrectly merges two different floors due to their similar appearance. Also, the sequence appears weak visual connectivity (less feature matches) at the stairs part, which cause an unstable camera pose estimation. On the other hand, our proposed method provides an accurate reconstruction with aligned structures of different floors.
    }
    \label{fig:corridor_sample}
\end{figure}
}

On the other hand, thanks to the progress of sensor technology, imaging devices equipped with other built-in sensors such as inertial measurement unit (IMU), become widely available~\cite{kelly2011visual,mourikis2007multi}. In the field of robotics perception, various visual-inertial odometry (VIO) algorithms~\cite{mourikis2007multi,jones2011visual,leutenegger2013keyframe,qin2018vins} have been proposed to provide an accurate local camera pose estimation by fusing IMU measurements to image information.
Even when the images cannot provide information about camera motion, VIO can still estimate the camera motion properly in a short time solely dependent on IMU measurements~\cite{qin2018vins}. 
Estimated camera poses, however, do not necessarily satisfy consistencies in the whole scene since online VIO systems rarely perform global bundle adjustment.
Thus, due to noises of IMU measurements, VIO often suffers from significant accumulated odometry errors.
In addition, because of its purpose, VIO only produces a rather small 3D map for each frame
than a globally consistent 3D structure obtained via SfM.

\begin{figure*}[t]
    \centering
    \includegraphics[width=1\textwidth]{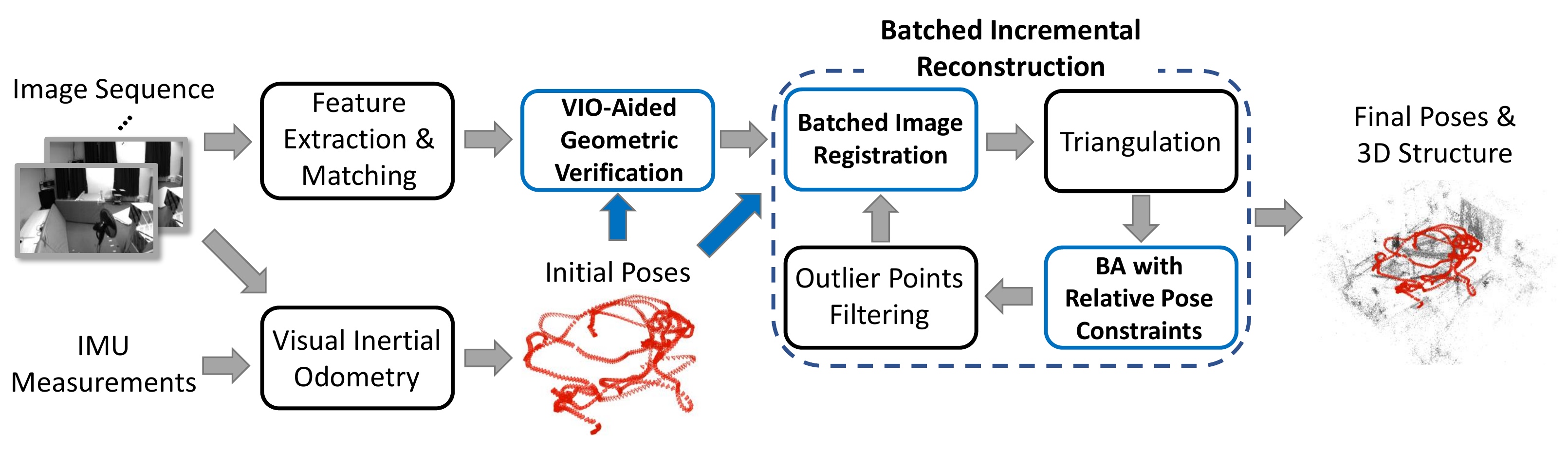}
    \caption{{\bf The overview of our proposed VIO-aided SfM system.} Our system takes sequential image collections and associated IMU measurements as inputs. We first obtain the initial camera poses of each image through a VIO system, and then incorporate them as prior information into the subsequent geometric verification and batched incremental reconstruction processes.}
    \label{fig:pipeline}
\end{figure*}

In this paper, we aim to achieve a robust and accurate 3D reconstruction that can produce a globally consistent 3D model. 
Assuming an input of sequential images and IMU measurements, we propose an SfM-based reconstruction pipeline that incorporates VIO estimation (illustrated in Fig.~\ref{fig:pipeline}). 
Exploiting its robustness and local consistency, our system firstly estimates the camera odometry via a VIO algorithm and then integrates the relative camera poses into each step of the SfM pipeline. 
This allows us to robustly construct 3D scene representations even in visually severe situations. 
Furthermore, our batch-wise image registration scheme with a new global bundle adjustment process, also aided by VIO estimation, ensures the global consistency of the obtained model at the marginal computational overhead. 
Fig.~\ref{fig:corridor_sample} provides a bird view sample of the 3D scene model obtained by our proposed method in a challenging scenario. 
Compared to an existing SfM-based reconstruction system~\cite{schonberger2016structure}, our method can produce an accurate 3D reconstruction, while dealing with repetitive scene natures and image sequences which has poor visual connectivity to each neighbor. 

Our contributions can be divided into three components:

\noindent
{\bf (1)} We propose a new geometric verification method that discards wrongly matched image pairs using a prior of geometric configuration from VIO. This scheme is especially effective in the presence of dominant repetitive structures in the scene.

\noindent
{\bf (2)} Each image frame in the input is incrementally registered to the model by initializing its pose using VIO estimation. We then introduce a new cost function of bundle adjustment that refines the camera poses and 3D structures while balancing the vision-based and VIO-based penalties.
Also, we effectively manage the computational cost for incrementally running the global bundle adjustment by designing the reconstruction pipeline in a batch-wise manner, while preserving the accuracy of the model.

\noindent
{\bf (3)} Finally, we evaluate the performance of the proposed pipeline using publicly available image (and IMU measurement) datasets which include various challenging situations such as weakly textured indoor scenes, industrial scenes dominated by repetitive structures, and poor lighting conditions.
Compared with SfM-based and VIO-based methods, the proposed method provides further accurate camera pose estimation, which results in a globally consistent 3D model. 

\section{Related Work \label{sec:related_work}}
\noindent
SfM has been widely used as a vision-based 3D reconstruction tool~\cite{wu2013towards,schonberger2016structure}, because of its robustness to various input scenarios~\cite{li2012worldwide,frahm2010building,aji2020embc} including un-ordered internet photos~\cite{snavely2006photo}. Several works achieved an accurate reconstruction for both camera motion and 3D structure, evolving each of sub-systems in SfM such as feature detection and matching~\cite{lowe2004distinctive,dong2015domain}, camera pose initialization~\cite{lepetit2009epnp,hesch2011direct,schonberger2016structure}, multi-view triangulation~\cite{hartley1997triangulation,hartley2003multiple} and bundle adjustment~\cite{triggs1999bundle,wu2013towards,kagami2020isie}. 
On the other hand, Simultaneous Localization and Mapping (SLAM) system has been developed as an alternative approach for vision-based 3D reconstruction, which is motivated to track the sequential image series input from a moving camera~\cite{davison2007monoslam,engel2014lsd,mur2015orb,forster2016svo,engel2017direct}. Ensuring its availability for real-time processing, SLAM usually employs on-the-fly consecutive camera poses estimation, and then refines them in post-processing, \eg, by loop closure~\cite{sunderhauf2012switchable, laskar2016robust, lajoie2019modeling}. 

One common shortage of these vision-based 3D reconstruction is the lack of relevance in the absence of visual information. 
Potential approach to address the issue is to collect auxiliary camera pose information from other systems such as Global Positioning System (GPS) and Inertial Navigation System (INS), and utilize them for the initial structure and camera poses~\cite{irschara2011efficient,aliakbarpour2015fast}. 
These methods still rely on high-precision GPS/INS measurements to provide a good global initialization. Cui~\etal try to eliminate the dependence using a track selection strategy and performing iterative triangulation and bundle adjustment~\cite{cui2015efficient}.

{\tabcolsep=5pt
\begin{figure*}[!ht]
    \centering
    \begin{tabular}{cc}
    \includegraphics[width=0.45\linewidth]{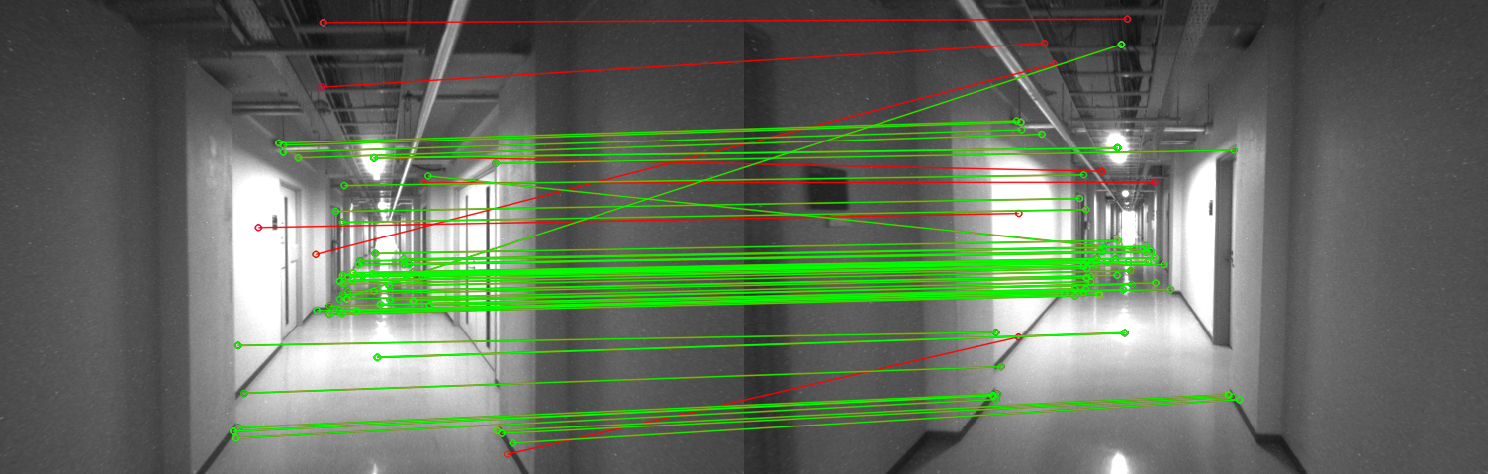} & 
    \includegraphics[width=0.45\linewidth]{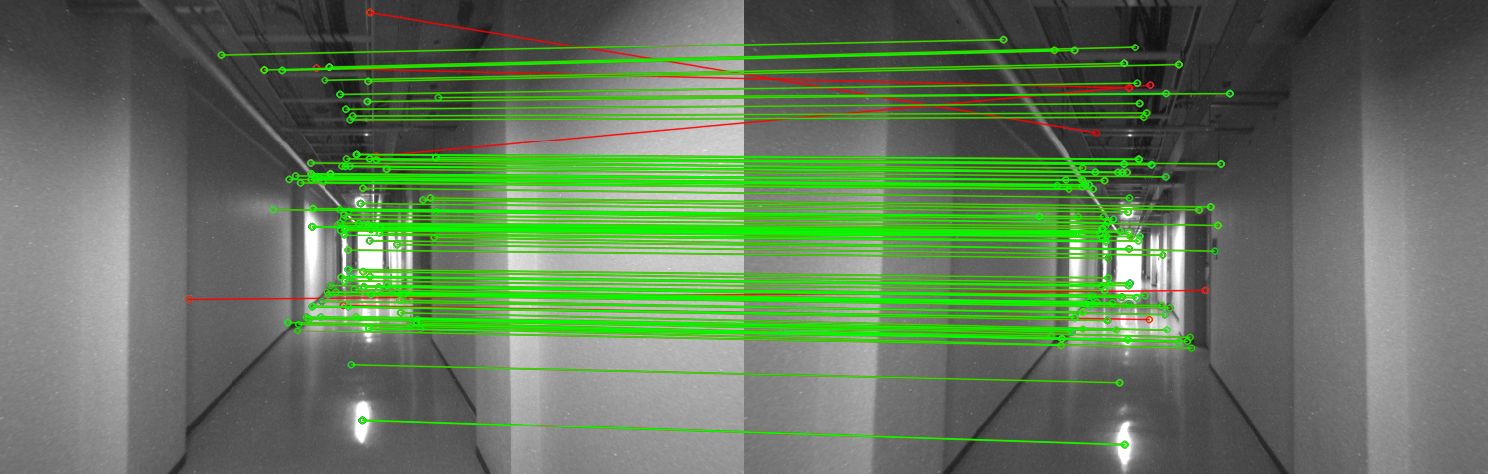} \\[-1pt]
    \bf{Wrong image pair taken on different floors} & \bf{Correct image pair taken on the same floor} \\[1pt]
    \multicolumn{2}{c}{(a) RANSAC-based geometric verification} \\[5pt]
    \includegraphics[width=0.45\linewidth]{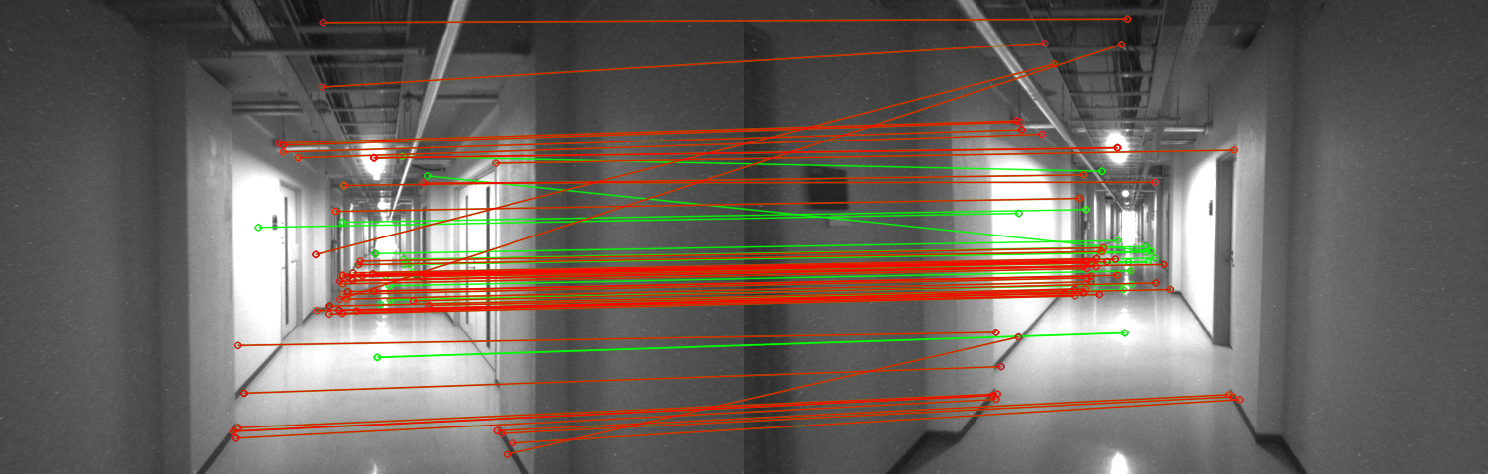} & 
    \includegraphics[width=0.45\linewidth]{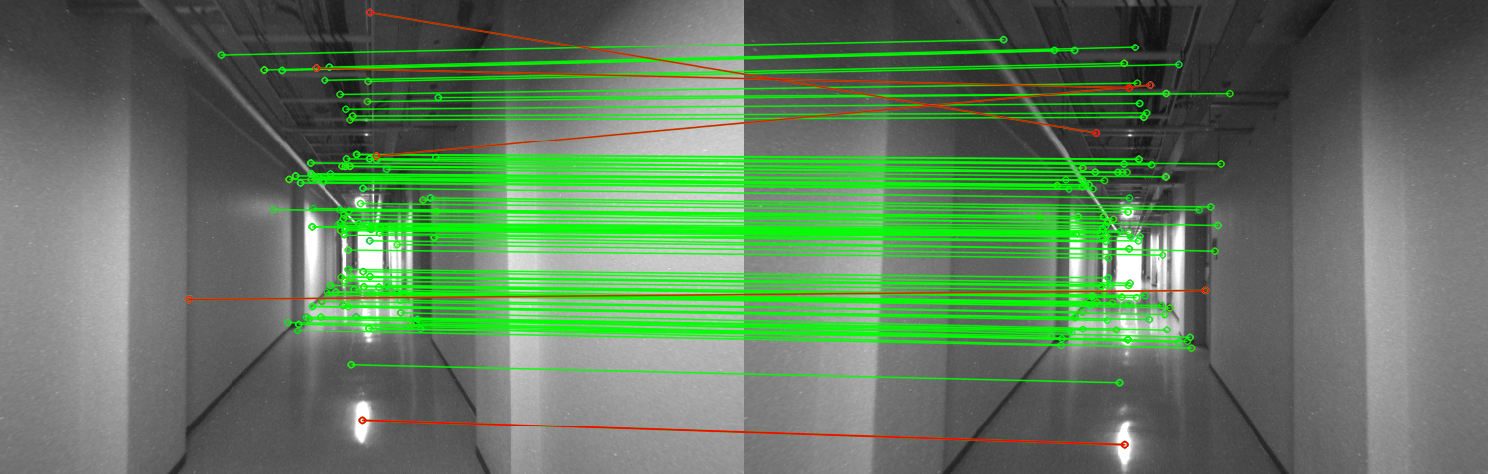} \\[-1pt]
    \bf{Wrong image pair taken on different floors.} & \bf{Correct image pair taken on the same floor.} \\[1pt]
    \multicolumn{2}{c}{(b) Our VIO-aided geometric verification}
    \end{tabular}
    \caption{{\bf Visualization of geometric verification under a challenging scene.} In the left column, we show a sample of a wrongly matched image pair, which looks very similar but comes from different places (taken on different floors). For comparison, the right column shows a sample of images taken on the same place.
    For each pair, we draw lines of matches by colors of green and red, indicating inlier and outlier matches detected by each geometric verification. 
    Our VIO-aided geometric verification evaluates image pairs by its outlier ratio (0.74 and 0.06 for the left and right pairs, respectively). 
    \label{fig:geometric_verification}
    }
\end{figure*}
}

In both computer vision and robotics perception, several VIO systems have been proposed to obtain more robust and accurate camera poses by fusing raw image information and IMU measurements in a single pose estimator~\cite{mourikis2007multi,jones2011visual,bloesch2017iterated,geneva2019efficient,leutenegger2013keyframe,qin2018vins}.
VIO systems usually seek the locally optimal cameras in a sliding-window fashion, \ie, considering only recent measurements~\cite{dong2012consistency,leutenegger2013keyframe,qin2018vins}.
This configuration, however, loses global information and sometimes causes a long-term drifting in estimated poses, \ie, cameras in the global coordinate system appears erroneous trajectories.
Several works address the drift issue via loop closure~\cite{konolige2008frameslam,mei2009constant,strasdat2010scale} that attempts to detect a camera trajectory loop during the camera motion. A global pose graph optimization~\cite{strasdat2010scale} is usually followed the loop detection to alleviate the drift errors.

In this paper, we aim to achieve high-quality reconstruction results in terms of both accuracy and robustness exploiting the camera poses obtained by a VIO system~\cite{qin2018vins}.
In contrast to previous methods~\cite{irschara2011efficient,aliakbarpour2015fast}, the obtained camera pose information is utilized in a batched incremental manner, which alleviates the impact of accumulated odometry errors. 
In addition, we introduce a constrained bundle adjustment using the camera motion from VIO to achieve a globally consistent and more robust reconstruction result. 

\section{3D Reconstruction by VIO-Aided SfM}
\noindent
Fig.~\ref{fig:pipeline} illustrates our proposed pipeline for sparse 3D reconstruction from sequential image collections and associated IMU measurements.
In what follows, we describe each part of our SfM-based reconstruction pipeline that incorporates the initial camera poses provided by a VIO system. 
First, we obtain camera poses of each image through an existing VIO system~\cite{qin2018vins}. 
Though the original system offers absolute camera poses in the scene, we extract relative camera poses for utilizing in latter processes.
Second, the camera poses are utilized as prior information for geometric verification of image pairs (Sec.~\ref{sec:geometric_verification}).
Third, we register images into the global 3D model in a batch-wise incremental manner that iteratively expands the model using local geometries of a subset of images (Sec.~\ref{sec:batched_reconstruction}).
In the end of each batched process, both obtained scene structure and registered camera poses are jointly optimized by also considering the VIO odometry to achieve both local and global consistency. The batch process is repeated until all images are registered.


\subsection{VIO-Aided Geometric Verification \label{sec:geometric_verification}}
\noindent
The accuracy of the 3D structure reconstructed by SfM highly depends on the detected correspondences between images. 
After getting tentative matches from local feature matching, SfM systems generally introduce an outlier rejection scheme such as RANSAC that fits a transformation model between image points computed from randomly sampled matches~\cite{fischler1981random, hartley2003multiple}.
However, as illustrated in Fig.~\ref{fig:geometric_verification} (a), an incorrect transformation can still be estimated when the dominant matches support a wrong model. 
This typically happens in weakly textured indoor scenes, which also include visually similar objects in different places, \eg, corridors, standardized doors, and furniture. 
Wrongly matched images offer false connections between actually distant places and can result in a collapsed 3D model (cf. Fig.~\ref{fig:corridor_sample}). 

We address such ambiguous matches by exploiting the robust camera pose estimation of the VIO system. 
Given an image pair $(I_t,I_{t'})$, VIO estimates the relative pose of $I_t$ with respect to $I_{t'}$, which consists of a rotation matrix ${\bf \hat{R}}$ and a translation vector ${\bf \hat{t}}$. 
Using this prior knowledge of camera motions, we introduce the pixel-based Epipolar Error (EE) that evaluates the local feature correspondences between images:
\begin{equation}
    \begin{split}
    &EE(f_{k,I_{t'}}, f_{k,I_t})=d(f_{k,I_{t'}},{\bf \hat{F}}f_{k,I_t}),\\
    &{\rm where} \;\; {\bf \hat{F}}={\bf K}^{-\mathrm{T}}{\bf \hat{t}}_\times{\bf \hat{R}}{\bf K}^{-1}.
    \end{split}
\end{equation}
$f_{k,I_{t}}$ and $f_{k,I_{t'}}$ are the corresponding feature points between the image pair, and {\bf K} is the camera intrinsic matrix. 
$[.]_\times$ denotes the matrix representation of the cross product and $d(.)$ denotes the perpendicular Euclidean distance between image point and line.
Feature correspondence $\{f_{k,I_{t}},f_{k,I_{t'}}\}$ is marked as outliers if the EE is larger than the threshold $T_{EE}$. 
Here ${\bf \hat{F}}$ is the fundamental matrix that projects the image point $f_{k,I_t}$ using known properties of the image intrinsic and the given prior camera poses. 
Consequently, EE evaluates the distances 
between image point $f_{k,I_{t'}}$ and the epipolar line ${\bf \hat{F}}f_{k,I_t}$, 
which presents the consistency between VIO-based geometric configuration and vision-based local correspondence. 
Therefore, the image pair giving a higher outlier ratio is more likely to build wrong visual matches. 
We assume the image pair is wrongly matched if 
the ratio of outliers exceeds $0.5$. 
 
Fig.~\ref{fig:geometric_verification} provides a typical sample on which EE can effectively help discarding a wrongly matched image pair. 
In the left sample, due to the visual similarity of the two different places captured in each of the images, most of the tentative matches support a false transformation model built via RANSAC~\cite{fischler1981random, hartley2003multiple}. 
In contrast, we build a transformation model using fairly correct camera poses obtained via VIO, thus can identify the wrong pair dominated by more than 50\% outliers.


Offering the new image-level verification strategy, we design a two-step geometric verification scheme.
We firstly build candidates of image pair for feature matching.
For each image, we choose $N_1$ neighboring images based on timestamp and $N_2$ visually similar images by image retrieval. 
The tentative correspondences between these pairs are obtained by standard feature matching scheme~\cite{lowe2004distinctive}.
As for the first stage, we verify each image pair based on its matches and a given prior fundamental matrix ${\bf \hat{F}}$. 
A wrong image pair detected based on the ratio of outliers is discarded and not anymore be used in latter processes. 
To get a more strictly verified set of feature correspondences, we next
evaluate the tentative matches of the remaining candidates by a RANSAC-based outlier rejection scheme~\cite{fischler1981random,hartley2003multiple} which estimates an epipolar transformation between images.
The candidates are finally approved if sufficient numbers of inlier matches exist.

\subsection{Batched Incremental Reconstruction \label{sec:batched_reconstruction}}
\noindent
We are next to register all images into the global coordinate system and build 3D scene points for recovering the target scene. Given initial camera poses obtained via VIO, the simplest strategy to achieve this goal is to register all images at once using absolute camera poses from VIO and triangulate 3D points using feature correspondences~\cite{irschara2011efficient,aliakbarpour2015fast}.
However, as mentioned in Sec.~\ref{sec:related_work}, VIO usually suffers from significant odometry drifts which turns into inaccurate absolute camera poses.
Instead, we utilize the relative camera pose of the image with respect to the previous frame and incrementally build up the model.
We also introduce to perform the image registration in a batched manner, which suppresses the additional computation costs in a marginal range.

\para{Batched Image Registration.}
We divide the sequential images into several consecutive $k$-frame batches in time order.
The batch size $k$ controls the final accuracy and computation time of our approach, which will be further discussed in Sec.~\ref{subsec:quantitative}.
The initial camera pose set of batch $i$ is denoted as ${\bf P}_i=\left\{{\bf p}\right\}$. In the $i$-th iteration, we register ${\bf P}_i$ by computing a rigid transformation ${\bf T}_i$ to align this batch of images to the current model. We directly compute ${\bf T}_i$ as the relative camera pose between the first image in ${\bf P}_i$ to the last image in ${\bf P}_{i-1}$, which proves to be fast and effective.
After each batched image registration, we triangulate new 3D scene points using feature matches between verified image pairs.

\para{Bundle Adjustment with Relative Pose Constraints.}
After each batched image registration and triangulation,
we refine the cameras and scene points to guarantee the global consistency of the reconstruction. 
One general scheme for this purpose is global bundle adjustment~\cite{triggs1999bundle} that minimizes the reprojection errors of the 3D scene points with respect to the estimated cameras and its observed feature points. 
We propose a new joint objective for bundle adjustment that relies not only on the visual information but also on the camera motion obtained via VIO: 
%
\begin{equation}
    \begin{split}
    \argmin_{{\bf p}_i,{\bf X}_j,{\bf K}} \sum_i\sum_j E_v({\bf p}_i,{\bf X}_j,{\bf K}) + \\
    \sum_i w_{i,i+1} E_r({\bf p}_{i},{\bf p}_{i+1} | {\bf \hat{p}}_{i,i+1}).
    \end{split}
\end{equation}
Here, $E_v$ is the standard bundle adjustment objective for SfM which is formulated as:
\begin{equation}
\label{eq:standard_BA}
  \begin{split}
  E_v({\bf p}_i,{\bf X}_j,{\bf K}) =
  \rho \left( \left \|{\bf z}_{ij}-\pi\left({\bf p}_i,{\bf X}_j,{\bf K}\right)\right\|^2 \right),
  \end{split}
\end{equation}
where ${\bf p}_i$ is the 6D absolute pose of the $i$-th camera  
and ${\bf z}_{i,j}$ is the feature point corresponding to the 3D point ${\bf X}_j$ observed by the $i$-th camera. 
$\pi(.)$ denotes the projection function which projects scene points to the image plane and $\rho$ is the robust function, \eg, Cauchy function. 
This objective, purely based on visual information, sometimes gives unstable results due to the weak visual connectivity between the scene points and images, \ie, lack of image features. 
We thus introduce a relative pose constraint term $E_r$ that penalizes the residuals of the camera motion with respect to the prior knowledge of the relative poses:
%
\begin{equation}
E_r({\bf p}_{i},{\bf p}_{i+1} | {\bf \hat{p}}_{i,i+1}) =
\left\|{\rm Log}({\bf p}_{i,i+1}, {\bf \hat{p}}_{i,i+1})\right\|^2,
\label{eq:baregularization}
\end{equation}
%
where ${\bf p}_{i,i+1}$ is the 6D relative camera pose computed by ${\bf p}_{i}$ and ${\bf p}_{i+1}$, 
and ${\bf \hat{p}}_{i,i+1}$ is the prior relative pose obtained from VIO.
${\rm Log}(.)$ denotes the log representation of the difference of two transformations.
Please notice that $E_r$ becomes a dominant term for determining $i$-th camera when the camera observes few 3D points, which penalizes the deviation between the estimated relative pose and its observation from VIO. 
Also considering the noisy nature of VIO estimation, we design the balancing factor $w_{i,i+1}$ as a self-adaptive function formulated by: 
\begin{equation}
w_{i,i+1}=\alpha e^{-\beta c_{i,i+1}}, 
\label{eq:lambda}
\end{equation}
where $c_{i,i+1}$ is the number of verified feature correspondences between two consecutive frames, and $\alpha$ and $\beta$ are hyper parameters. 
Consequently, the objective heavily depends on the vision-based penalty $E_v$ when there is sufficient visual connectivity between cameras. 
We use Ceres Solver~\cite{ceres-solver} for solving this nonlinear problem. After the bundle adjustment, the scene points with large projection errors or small triangulation angles are filtered out for further robustness.

\section{Experiments \label{sec:experiments}}
\begin{table*}[!ht]
  \centering
  \resizebox{\textwidth}{22mm}{
    \begin{tabular}{|c|c|c|c|c|c|c|c|c|c|c|c|c|c|}
    \hline
    \multicolumn{2}{|c|}{}                           & \multicolumn{6}{c|}{Visual methods}                                                                                                                  & \multicolumn{6}{c|}{Visual-inertial methods}                                                                              \\ \cline{3-14} 
    \multicolumn{2}{|c|}{\multirow{-2}{*}{Sequence}} & \multicolumn{2}{c|}{COLMAP\cite{schonberger2016structure}}                                 & \multicolumn{2}{c|}{ORB-SLAM2\cite{mur2015orb}}                              & \multicolumn{2}{c|}{DSO\cite{engel2017direct}} & \multicolumn{2}{c|}{OKVIS\cite{leutenegger2013keyframe}} & \multicolumn{2}{c|}{VINS-Mono\cite{qin2018vins}} & \multicolumn{2}{c|}{Ours}                                   \\ \hline
    Name                       & Frames & RMSE                         & ME                       & RMSE                         & ME                       & RMSE       & ME      & RMSE        & ME       & RMSE          & ME         & RMSE                         & ME                       \\ \hline
    V1\_02\_medium             & 756 & {\color[HTML]{3531FF} 0.043} & {\color[HTML]{3531FF} 0.040} & 0.064                        & 0.063                        & 0.598      & 0.213       & 0.067       & 0.062        & 0.060         & 0.057          & {\color[HTML]{FF0000} 0.022}               & {\color[HTML]{FF0000}0.019}               \\ \hline
    V1\_03\_difficult          & 916 & {\color[HTML]{3531FF} 0.054} & {\color[HTML]{3531FF} 0.051} & 0.531                        & 0.235                        & 0.925      & 0.935       & 0.105       & 0.089        & 0.173         & 0.131          & {\color[HTML]{FF0000}0.043}               & {\color[HTML]{FF0000}0.032}               \\ \hline
    V2\_02\_medium             & 1025 & {\color[HTML]{3531FF} 0.029} & {\color[HTML]{3531FF} 0.032} & 0.056                        & 0.056                        & 0.092      & 0.080       & 0.081       & 0.058        & 0.124         & 0.103          & {\color[HTML]{FF0000}0.014}               & {\color[HTML]{FF0000}0.012}               \\ \hline
    V2\_03\_difficult          & 1028 & \begin{tabular}[c]{@{}c@{}}{\color[HTML]{3531FF} 0.041}\\ (1014)\end{tabular} & \begin{tabular}[c]{@{}c@{}}{\color[HTML]{3531FF} 0.036}\\ (1014)\end{tabular} & \begin{tabular}[c]{@{}c@{}}0.079\\ (900)\end{tabular}                        & \begin{tabular}[c]{@{}c@{}}0.073\\ (900)\end{tabular}                        & 1.386      & 1.008       & -       & -        & 0.191         & 0.153          & {\color[HTML]{FF0000}0.029}               & {\color[HTML]{FF0000}0.021}               \\ \hline
    MH\_03\_medium             & 968 & {\color[HTML]{000000} 0.040} & {\color[HTML]{000000} 0.034} & {\color[HTML]{3531FF} 0.038} & {\color[HTML]{3531FF} 0.032} & 0.172      & 0.135       & 0.146       & 0.143        & 0.080         & 0.067          & {\color[HTML]{FF0000}0.035}               & {\color[HTML]{FF0000}0.029}               \\ \hline
    MH\_04\_difficult          & 681 & 0.095                        & 0.078                        & {\color[HTML]{FF0000}0.059}               & {\color[HTML]{FF0000}0.049}               & 0.172      & 0.171       & 0.138       & 0.131        & 0.124         & 0.123          & {\color[HTML]{3531FF} 0.092} & {\color[HTML]{3531FF} 0.077} \\ \hline
    MH\_05\_difficult          & 701 & 0.084                        & {\color[HTML]{3531FF} 0.064} & {\color[HTML]{FF0000}0.068}               & {\color[HTML]{FF0000}0.055}               & 0.102      & 0.093       & 0.261       & 0.227        & 0.133         & 0.110          & {\color[HTML]{3531FF} 0.083} & 0.072                        \\ \hline
    \end{tabular}
    }
  \caption{{\bf Performance of reconstructed trajectory.} 
  For each comparison in the column, we report the root mean square error (RMSE) and the median error (ME) of the reconstructed camera positions in meters\protect\footnotemark. 
  Best results are presented in bold and the second best are in blue. We additionally report the number of reconstructed cameras where the method does not provide a full trajectory for the input sequences.
}
  \label{table:euroc}
\end{table*}
\para{Datasets.} We collect several challenging sequences under different environments from two publicly available datasets~\cite{burri2016euroc,Kasper-IROS-2019};
{\bf EuRoC Dataset}~\cite{burri2016euroc} contains sequences of images in 20Hz and IMU measurements in 200Hz captured mainly in indoor scenes. 
It also provides a ground-truth camera pose of each image which is obtained via VICON and Leica MS50. 
Each sequence is labeled as {\it easy}, {\it medium}, or {\it difficult} according to the illumination and camera motion. 
We use seven sequences labeled with {\it medium} and {\it difficult} for evaluation, which capture separated scenes shown in Fig.~\ref{fig:samples};
Similarly, {\bf OIVIO Dataset}~\cite{Kasper-IROS-2019} consists of 36 sequences of images in 30Hz and IMUs in 200Hz recorded in weakly lighted environments. Since the dataset does not provide ground-truth poses for all images, we only report qualitative samples of reconstruction results for two selected sequences named as ``TUNNEL HANDHELD 3'' and ``MINE HANDHELD 1'' (denoted by ``Tunnel'' and ``Mine'') in Fig.~\ref{fig:qualitative}. 
For both datasets, we arrange image sequences by removing the static frames at the beginning and the end, and downsampling to 10Hz. 
%

\para{Implementation.} 
We construct our reconstruction system based on COLMAP, an existing SfM tool~\cite{schonberger2016structure} implemented in C++. 
$N_1$ and $N_2$ controlling the number of image pairs to match, is set to 40 and 30, which are reasonably fit to the frequency of the input image sequence (10Hz). 
Also considering errors of VIO estimation, we empirically set a relatively loose threshold of EE for our VIO-aided geometric verification: $T_{EE}=20$ pixels.
Batch size $k$ controlling the reliability of our incremental SfM onto the initial VIO estimation is set to 50 frames. We also report a validation of $k$ in Sec.~\ref{subsec:quantitative}. 
The hyper parameters $\alpha$ and $\beta$ in Eq.~\eqref{eq:lambda} are set to 1e3 and 0.003.
All experiments are conducted on a desktop PC with an Intel Core i7-6700 CPU and a Geforce GTX 980Ti GPU.

\subsection{Quantitative Evaluation for Reconstructed Odometry \label{subsec:quantitative}}
\noindent
We first evaluate the accuracy of the reconstructed cameras on the 7 sequences from EuRoC dataset where the camera location ground truth is available. 
We compare our approach with other visual and visual-inertial methods including
COLMAP~\cite{schonberger2016structure}, ORB-SLAM2~\cite{mur2015orb}, DSO~\cite{engel2017direct}, OKVIS~\cite{leutenegger2013keyframe} and VINS-Mono~\cite{qin2018vins}, using implementations provided by authors. 
Before evaluation, we compute a similarity transformation~\cite{umeyama1991least} to align the estimated trajectory to the ground-truth.

Tab.~\ref{table:euroc} reports the quantitative comparisons of reconstructed trajectories.
Our method produces smaller errors than other existing visual-inertial methods, which locally estimate the camera motion, and shows the best performance in all comparisons on five sequences. 
Typical improvements can be seen in the first four sequences which include many aggressive and complex movements. 
Vision-based SfM (COLMAP) shows the second-best performances on these sequences, but fails to estimate a full trajectory in V2\_03\_difficult (missing 14 frames highlighted in Fig.~\ref{fig:RPC} (a)) due to strong motion blur and texture-less areas in the sequence. 
On the other hand, our batched image registration succeeds to register all images by using the initial trajectory of each batched section provided by VIO.
%
%
{\tabcolsep=5pt
\begin{figure}[t]
    \centering
    \begin{tabular}{cc}
    \includegraphics[width=0.45\linewidth]{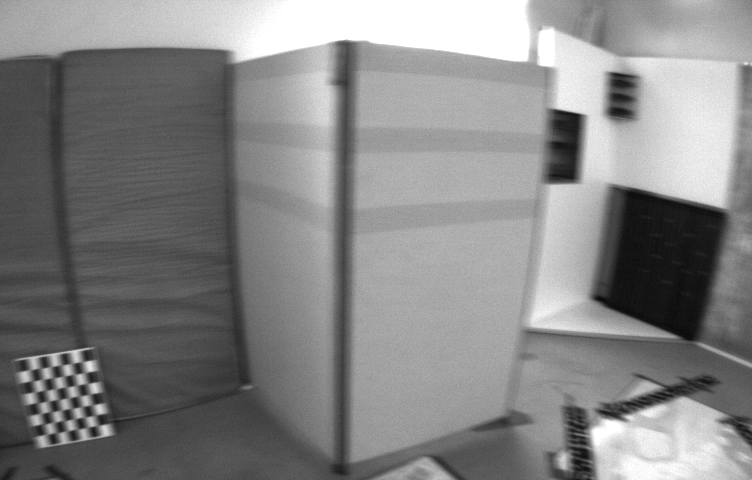} & 
    \includegraphics[width=0.45\linewidth]{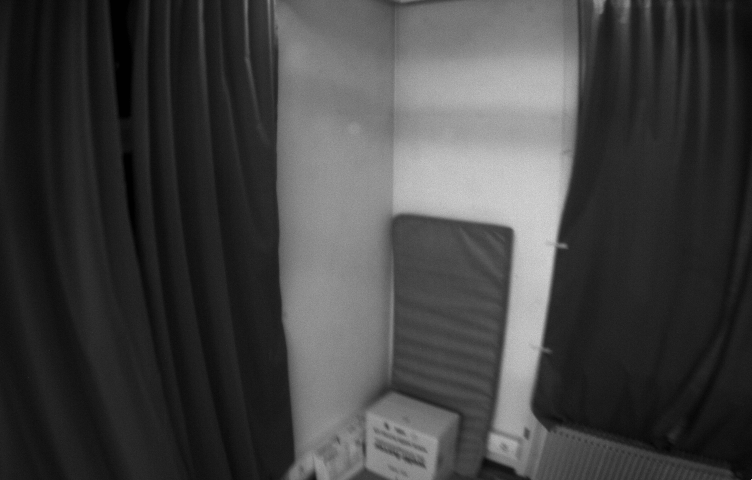} \\[-3pt]
    (a) V1\_03\_difficult & (b) V2\_03\_difficult \\[1pt]
    \includegraphics[width=0.45\linewidth]{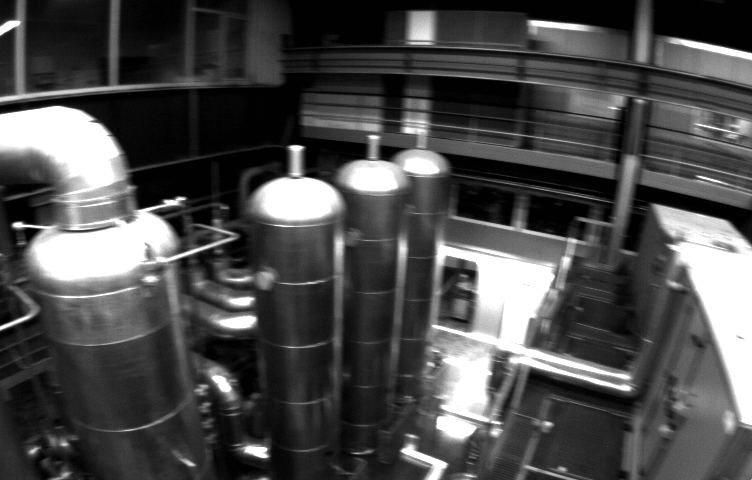} & 
    \includegraphics[width=0.45\linewidth]{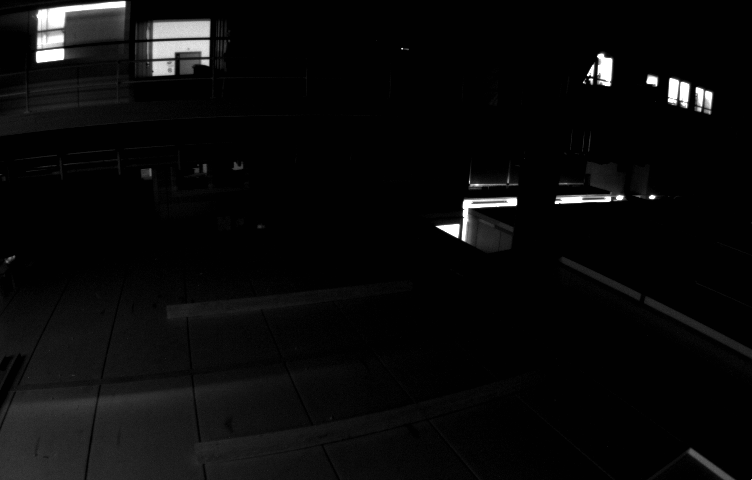} \\[-3pt]
    (c) MH\_04\_difficult & (d) MH\_05\_difficult
    \end{tabular}
    \caption{Sample images for the challenging environments in the EuRoC Dataset~\cite{burri2016euroc}.
    }
    \label{fig:samples}
\end{figure}
}

{\tabcolsep=1pt
\begin{figure*}[ht]
\centering

\begin{tabular}{cccc}
    \includegraphics[height=4.8cm]{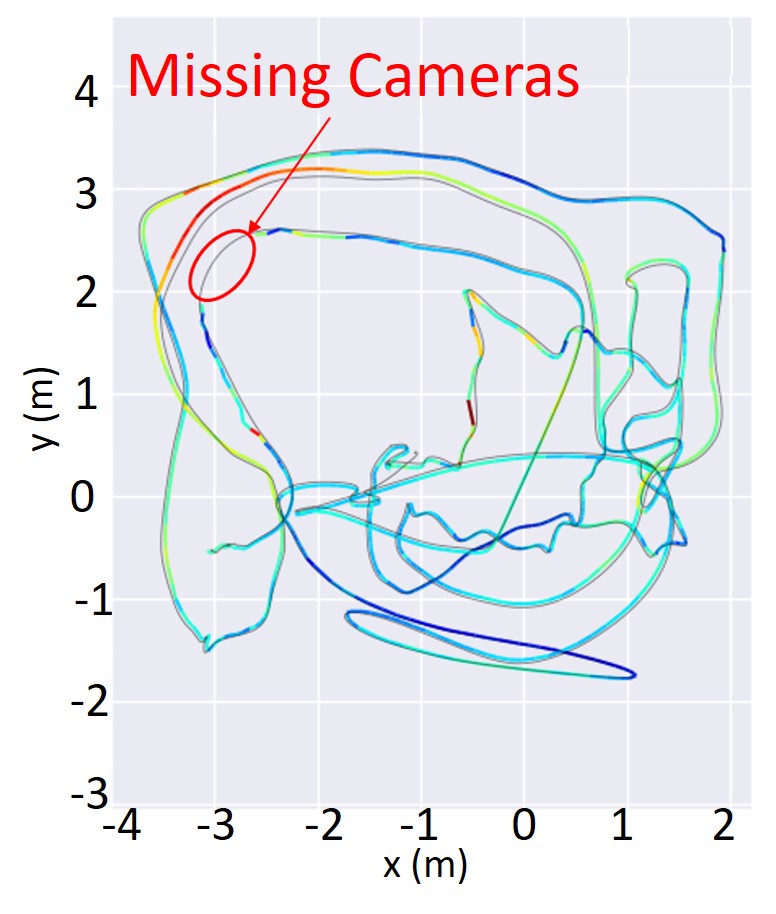}  & 
    \includegraphics[height=4.8cm]{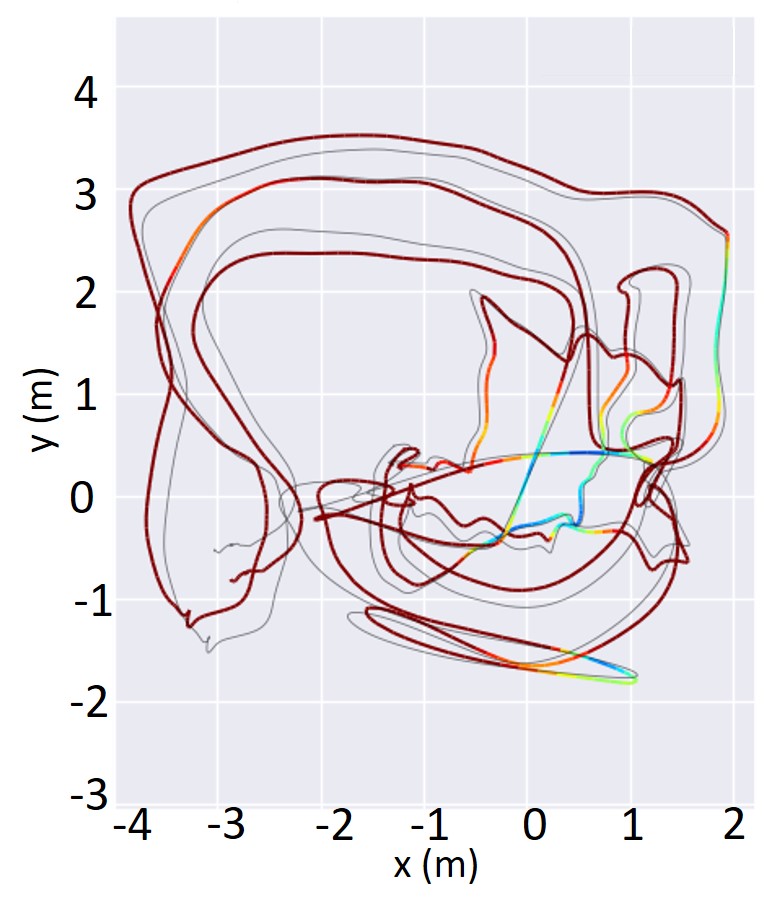} & 
    \includegraphics[height=4.8cm]{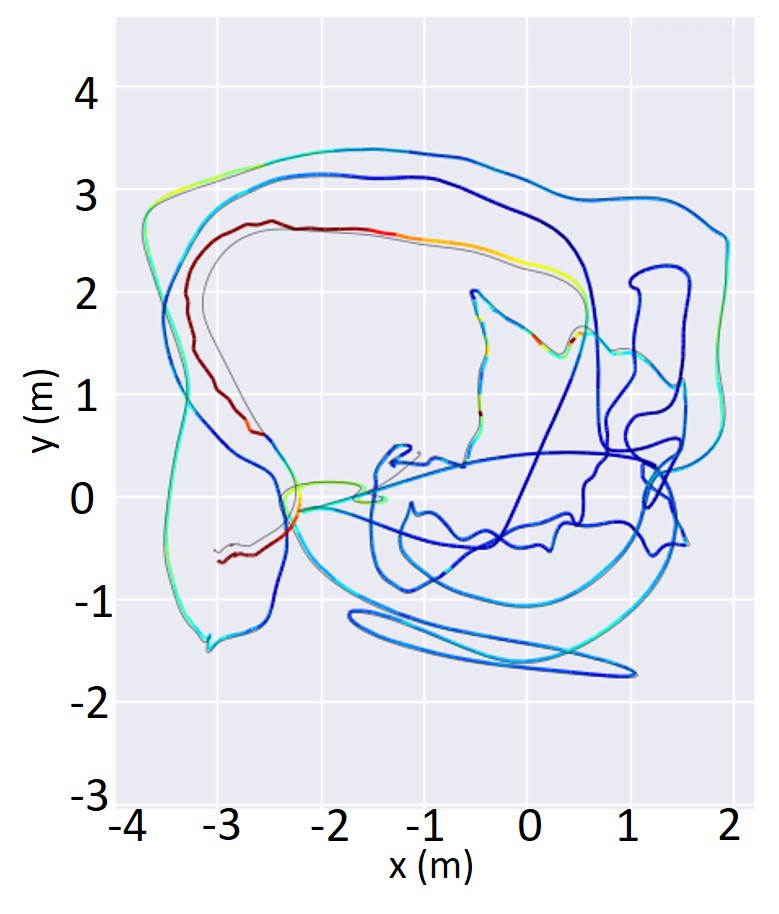} & 
    \includegraphics[height=4.8cm]{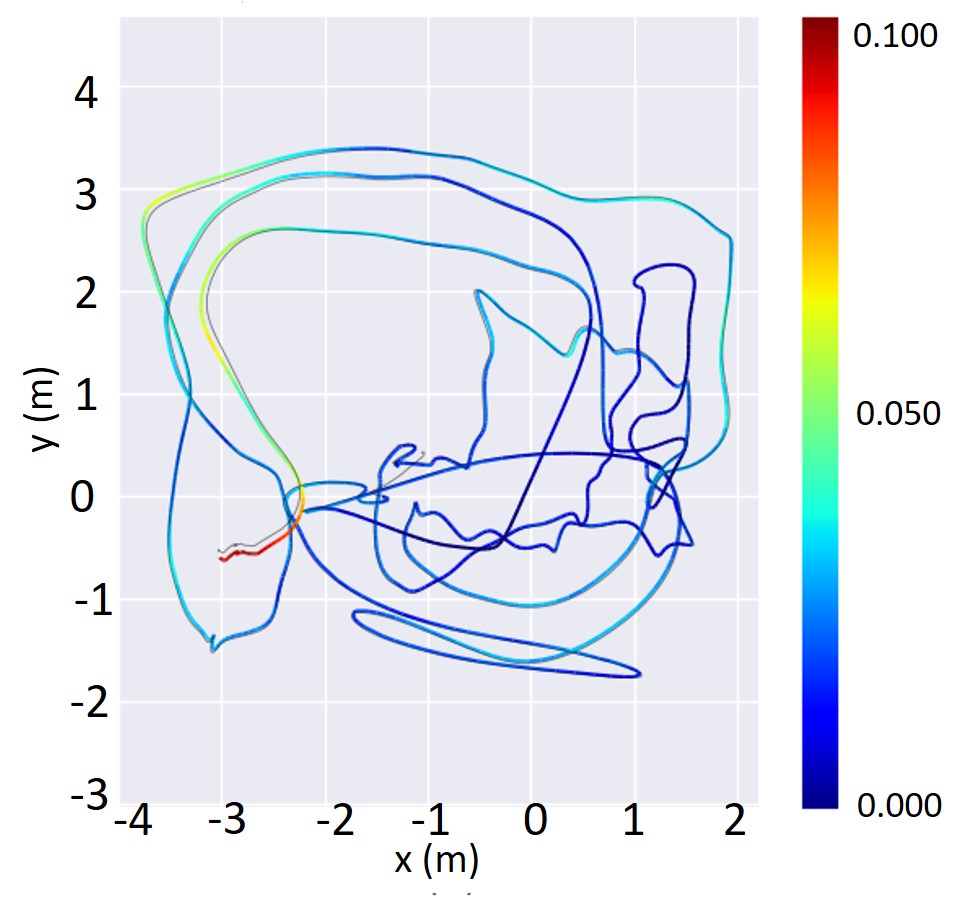} \\[-3pt]
    (a) COLMAP~\cite{schonberger2016structure} &
    (b) VINS-Mono~\cite{qin2018vins} &
    (c) Proposed method w/o RPC &
    (d) {\bf Proposed method} \\[-3pt]
\end{tabular}

\caption{{\bf Recovered camera trajectories on V2\_03\_difficult sequence from EuRoC Dataset~\cite{burri2016euroc}.}
The estimated camera trajectories are presented as line segments colored by the absolute position error (meters) with respect to the ground-truth camera trajectory (gray line). 
'w/o RPC' means using a standard BA without relative pose constraint in the proposed method. 
RMSE of each trajectory is (a) 0.041, (b) 0.191, (c) 0.043 meters, (d) 0.029 meters, respectively. 
}
\label{fig:RPC}
\end{figure*}
}


\begin{figure}[ht]
\centering
\includegraphics[width=1\linewidth]{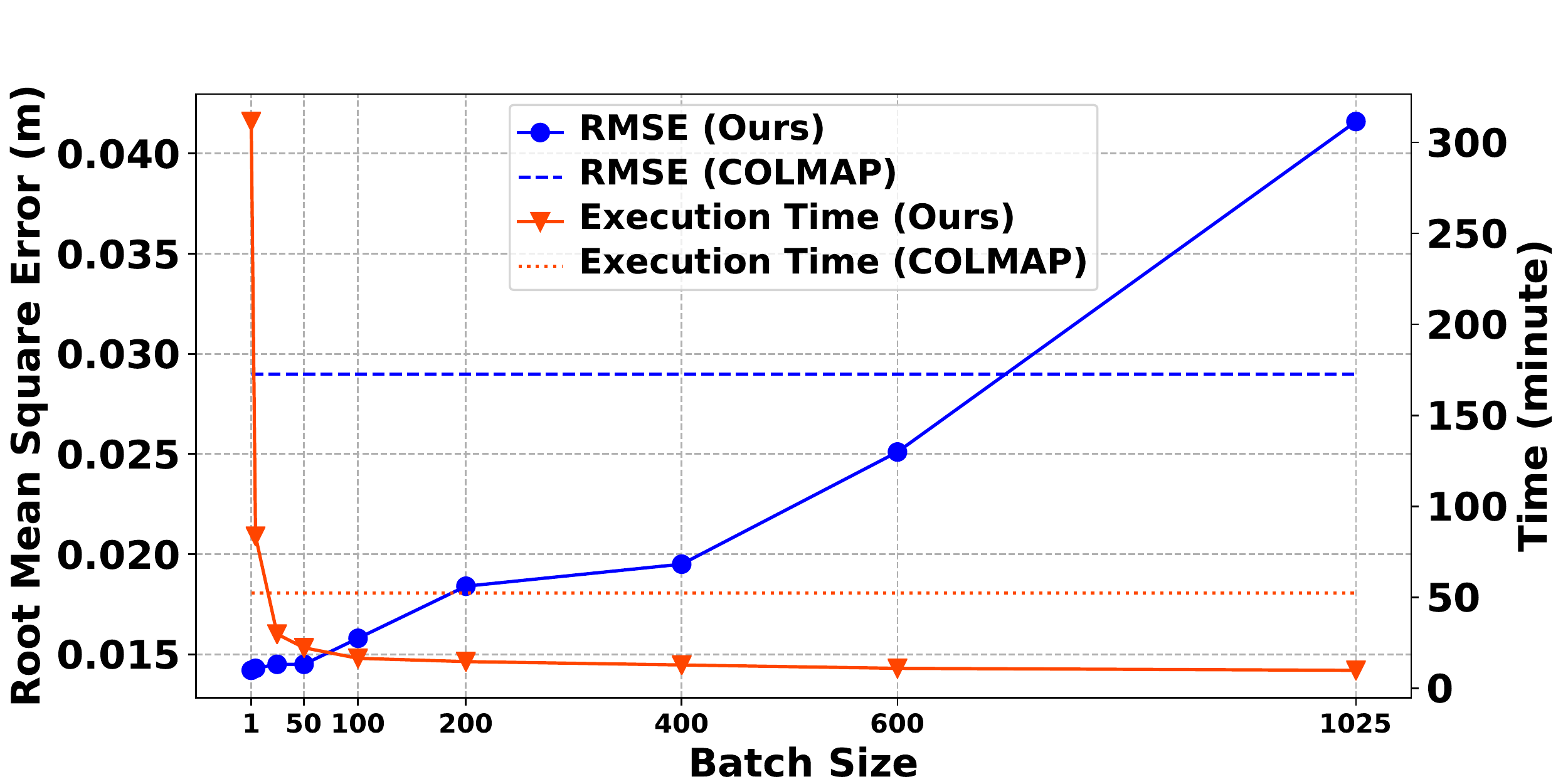}
\caption{{\bf Trajectory RMSE and execution time with respect to the batch size.} The trajectory RMSE and the execution time of COLMAP~\cite{schonberger2016structure} are also shown by dotted lines for comparison.}
\label{fig:batch_2}
\end{figure}

Fig.~\ref{fig:RPC} shows the camera trajectories estimated by our method and others consisting each component of our full pipeline.
Our SfM-based method (d) generates a highly accurate and complete trajectory using both visual and inertial information, whereas purely vision-based SfM (a) fails to track all cameras. 
Please also notice that VINS-Mono (b), which provides the initial camera motions in our method, gives a globally inconsistent trajectory, and our method shows a remarkable boost in accuracy. This fact demonstrates that our batched image registration can effectively deal with accumulated drifts of the initial camera poses by dividing sequences and using camera motions only in the local section. 
We also construct a variant of our method (c) that employs a standard BA ignoring the relative pose constraint (Eq.~\eqref{eq:baregularization}). 
This variant results in a locally erroneous camera trajectory compared to our full pipeline (d), suffering from the lack of proof of camera poses when there is less visual connectivity to neighboring images. 

\para{Influence of Batch Size. }
Second, we analyze the influence of batch size $k$ by evaluating our method with varying the parameter. 
The experiment is carried out on the V2\_02\_medium sequence from the EuRoC dataset, which contains 1025 images. 
Fig.~\ref{fig:batch_2} shows the RMSE of the recovered camera trajectory and execution time of reconstruction with various settings of the batch size. 
In general, a smaller batch size can bring a higher accuracy, but at the cost of computation time since the number of bundle adjustment increases. We find that when the batch size is smaller than a certain value, the accuracy does not change significantly.
On the other hand, the execution time increases tremendously when the batch size is extremely small.
This can be attributed to the computational costs for bundle adjustment which runs in each batched reconstruction process. 
The smaller batch size leads the more iteration number and consequently produces further complexities in latter iterations which refine a large number of registered cameras and points.
Considering the trade-off between accuracy and efficiency, we finally select 50 as the batch size in our experiments.

\footnotetext{We do not report results of OKVIS on V2\_03\_difficult because despite our best efforts, it fails to estimate a reasonable trajectory for the sequence.}
\subsection{Qualitative Evaluation for Reconstructed 3D Models \label{subsec:qualitative}}

\begin{figure*}[!ht]
\centering
\begin{tabular}{c}
    \includegraphics[width=0.9\textwidth]{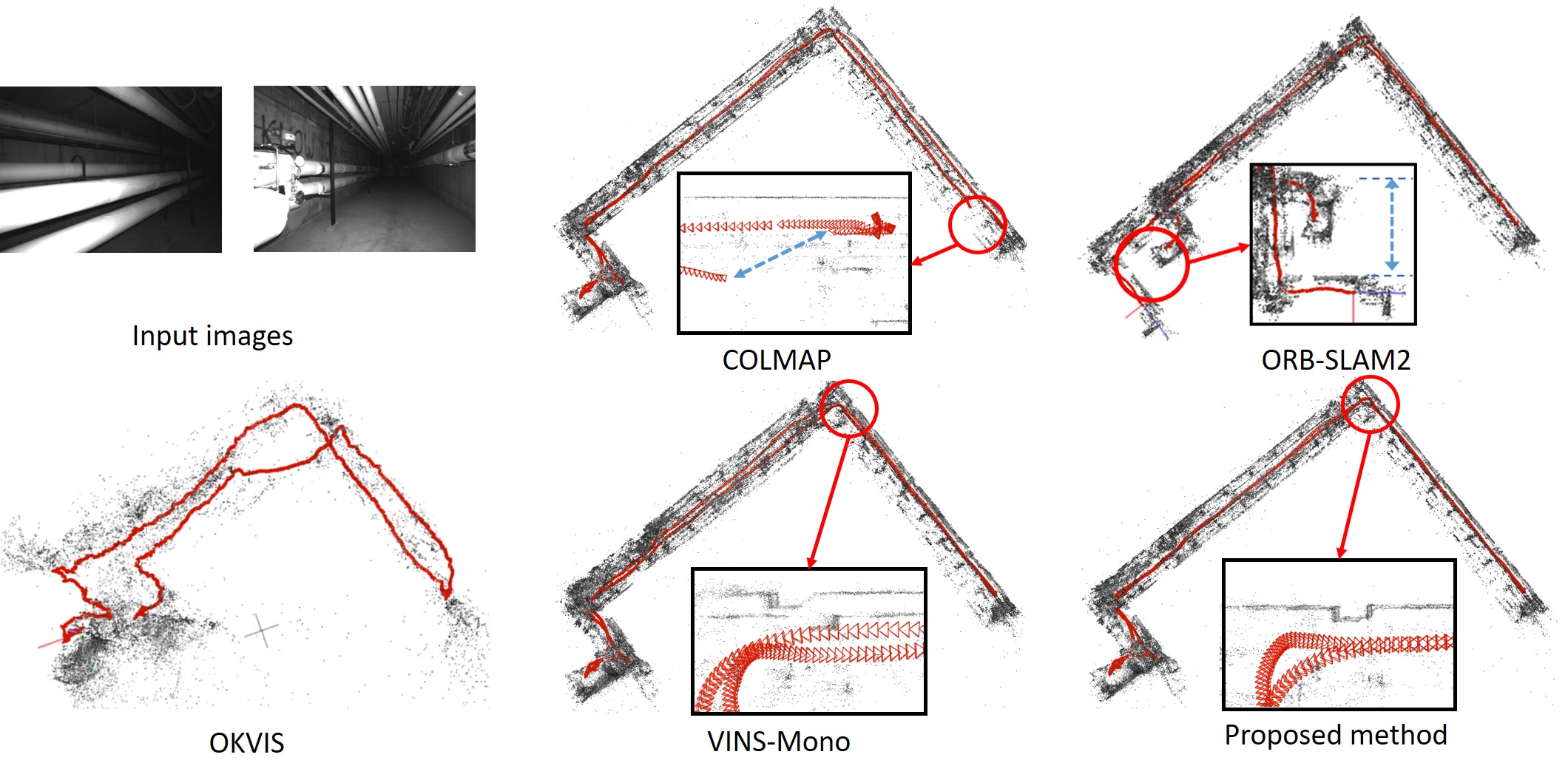} \\[-3pt]
    (a) Tunnel sequence \\[3pt]
    \includegraphics[width=0.9\textwidth]{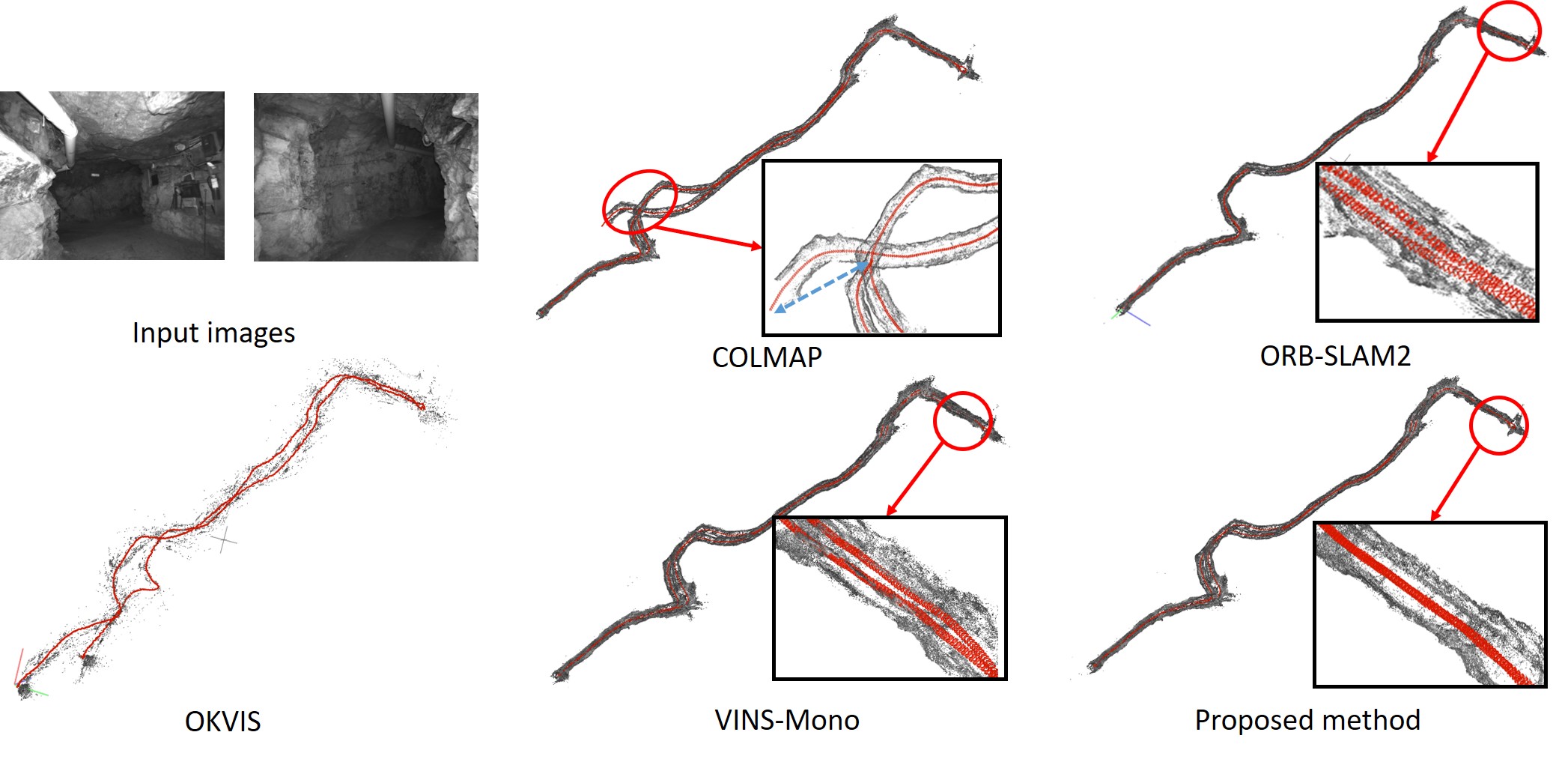} \\[-3pt]
    (b) Mine sequence \\[-3pt]
\end{tabular}
\caption{{\bf Reconstruction results for two challenging environments: Tunnel and Mine from OIVIO dataset~\cite{Kasper-IROS-2019}}. We compare our proposed method with COLMAP~\cite{schonberger2016structure}, ORB-SLAM2~\cite{mur2015orb}, OKVIS~\cite{leutenegger2013keyframe} and VINS-Mono~\cite{qin2018vins}.
}
\label{fig:qualitative}
\end{figure*}

\noindent
We finally provide some visual samples of the reconstructed cameras and 3D points for challenging image sequences from the OIVIO dataset~\cite{Kasper-IROS-2019}.
Fig.~\ref{fig:qualitative} shows the reconstruction results obtained by our method, COLMAP~\cite{schonberger2016structure}, ORB-SLAM2~\cite{mur2015orb}, OKVIS~\cite{leutenegger2013keyframe} and VINS-Mono~\cite{qin2018vins}. 
Since OKVIS~\cite{leutenegger2013keyframe} and VINS-Mono~\cite{qin2018vins} do not output 3D points, we additionally obtain 3D points for them by triangulating feature tracks approved by our geometric verification.
We exclude DSO~\cite{engel2017direct} from this comparison because it fails to estimate complete trajectories for these sequences. 
For both of the scenes, COLMAP~\cite{schonberger2016structure} fails to construct the continuous camera trajectories (pointed by blue arrows) due to the weak visual connectivity of the sequences. This especially happens in the Mine sequence, which has a longer trajectory, resulting in an inconsistent 3D model with duplicated structures (Fig.~\ref{fig:qualitative}~(b)).
On the other hand, we build a continuous camera trajectory by initializing camera poses using VIO estimation.
Please also note that other SLAM or visual-inertial methods, including ORB-SLAM2~\cite{mur2015orb}, OKVIS~\cite{leutenegger2013keyframe} and VINS-Mono~\cite{qin2018vins}, suffer from inconsistent model reconstruction, \eg, drifted and duplicated structures, because they estimate a locally consistent camera trajectory mainly based on feature matching only against to neighboring frames.
In contrast, our system provides globally consistent 3D models by employing global bundle adjustment after registering each batched trajectory. 

To sum up, we achieve the globally consistent 3D reconstruction that introduces the camera trajectory from VIO to an SfM pipeline. Experiments demonstrate that our method deals with challenging scenes providing less visual information while effectively utilizing the prior knowledge of camera poses from a locally consistent VIO estimation. 

\section{Conclusion}
\noindent
In this paper, we have proposed a SfM-based 3D reconstruction pipeline that effectively takes advantage of the camera pose information from a VIO.
In contrast to existing SLAM-based visual-inertial reconstruction methods, we aim to construct a globally consistent and complete 3D model including camera poses and 3D points. 
Our method consists of a simple combination of VIO-aided camera poses initialization and SfM-based images-points reconstruction, still gives a great margin in terms of accuracy of the reconstructed model. 
Experiments on publicly available datasets demonstrate that our system can achieve an accurate and robust 3D reconstruction in challenging environments where the images provide less visual evidences for reconstruction. 
Also, we have shown that the computational time for the reconstruction can effectively be reduced by a batched incremental reconstruction process.
One of the future works would be the detailed analysis for determining several parameters in our system that are highly relevant to its feasibility to noisy IMUs and consequent erroneous VIO estimation. Although we are currently tuning these parameters empirically, we believe that this work still suggests that the prior knowledge of camera motion can benefit existing vision-based 3D reconstruction systems and implies there is significant room for improvement towards accurate 3D reconstruction.

\noindent
{\bf Acknowledgement}. This work is partly supported by JSPS KAKENHI Grant Number 17H00744.

{\small
\bibliographystyle{IEEEtran}
\bibliography{shortstrings,egbib}
}

\end{document}